# On the Integration of Optical Flow and Action Recognition

Laura Sevilla-Lara
Facebook
laurasevilla@fb.com

Yiyi Liao
Zhejiang University
yiyi.liao@tuebingen.mpg.de

Fatma Güney
MPI for Intelligent Systems
fatma.guney@tuebingen.mpg.de

Varun Jampani
NVIDIA
varunjampani@gmail.com

Andreas Geiger
MPI for Intelligent Systems
andreas.geiger@tuebingen.mpg.de

Michael J. Black
MPI for Intelligent Systems
black@tuebingen.mpg.de

## Abstract

*Most of the top performing action recognition methods use optical flow as a "black box" input. Here we take a deeper look at the combination of flow and action recognition, and investigate why optical flow is helpful, what makes a flow method good for action recognition, and how we can make it better. In particular, we investigate the impact of different flow algorithms and input transformations to better understand how these affect a state-of-the-art action recognition method. Furthermore, we fine tune two neural-network flow methods end-to-end on the most widely used action recognition dataset (UCF101). Based on these experiments, we make the following five observations: 1) optical flow is useful for action recognition because it is invariant to appearance, 2) optical flow methods are optimized to minimize end-point-error (EPE), but the EPE of current methods is not well correlated with action recognition performance, 3) for the flow methods tested, accuracy at boundaries and at small displacements is most correlated with action recognition performance, 4) training optical flow to minimize classification error instead of minimizing EPE improves recognition performance, and 5) optical flow learned for the task of action recognition differs from traditional optical flow especially inside the human body and at the boundary of the body. These observations may encourage optical flow researchers to look beyond EPE as a goal and guide action recognition researchers to seek better motion cues, leading to a tighter integration of the optical flow and action recognition communities.*

## 1. Introduction

Traditionally, the computer vision problem has been divided into sub-problems or modules that are easier to solve, with the goal of putting them together later to solve the larger problem. Some examples include motion, depth, and edge estimation at the lower level, attribute extraction at the middle level, and object or action recognition at the higher level. This had yielded rapid progress in separate communities focused on specific sub-problems.

In the video domain, two common sub-problems are motion estimation and action recognition, which are most commonly combined in a straightforward way. Motion is estimated as the optical flow of the scene, typically by minimizing brightness constancy error, with the goal of achieving accurate flow in terms of end-point-error (EPE), which is the average Euclidean distance between the ground truth and the estimated flow. The optical flow and the raw images are the input to the next module, which computes the video label [28, 34]. This approach, although widely used, is based on a few untested hypotheses that we examine in this paper.

**Hypothesis 1:** *The optical flow between two frames is a good feature for video classification.* While it may seem intuitive to include image motion in a task related to video, often video categories in current datasets can be identified from a single image as illustrated in Fig. 1. Many actions are uniquely defined by human-object interactions, which can be recognized in a single frame [36]. Thus one may ask whether motion, in the form of optical flow, is necessary for action recognition. There is the observation in the literature [33], however, that classification accuracy using exclusively optical flow as input, can be even higher than classification accuracy using exclusively the raw images in some standard datasets [20, 30]. This is surprising since current datasets show a large correlation between scenes/objects and categories (*i.e.* water and the category "Kayaking", or guitars and "PlayingGuitar" appear together often), and flow seems to lack much of this explicit object information.

Consequently, we ask: *What makes motion such a useful feature for video classification?* Intuitively it would seem that motion trajectories contain information useful for recognition. We test this hypothesis by randomly shuffling the flow fields in time. We observe that the accuracy decreases only a small amount, suggesting that trajectories are not the main source of information in optical flow. Since the flow has been computed between adjacent frames, there is still temporal information in the input. Consequently, we remove the temporal information completely by shuffling the frames *before* computing the flow. We observe that the accuracy decreases further but is still 60 times higher than chance. Thus we argue that most of the value of optical flow in current architectures is that it is a representation of the scene invariant to appearance. This may be related to recent findings [7] that show that at very large scale, networks using only optical flow have lower accuracy than networks using only images. The intuition is that as the training set becomes larger, more examples of different illuminations, clothing or backgrounds are seen, making generalization easier and invariance to appearance less crucial.

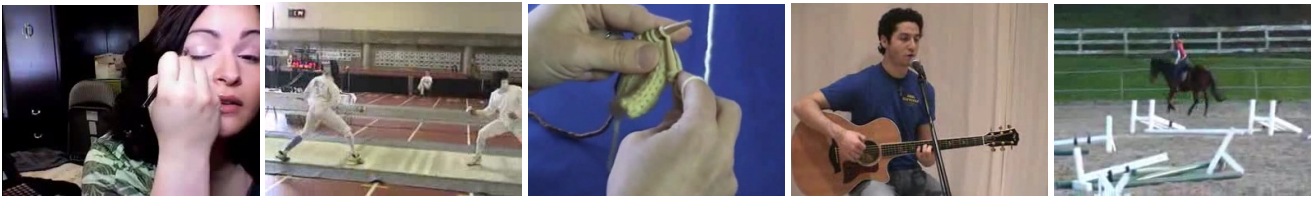

Figure 1. **Frames from the UCF101 dataset [30]**, sampled from the categories *ApplyEyeMakeup, Fencing, Knitting, PlayingGuitar, HorseRiding*. A single frame often contains enough information to correctly guess the category.

We test the hypothesis that much of the value of optical flow is that it is a representation invariant to appearance. For this we alter the appearance of the input and observe that accuracy from optical flow goes down marginally by 1%, while accuracy from raw images goes down by 50%. This suggests that the motion trajectories are not the root of the success of optical flow, and that establishing useful motion representations remains an open problem that optical flow on its own does not address.

**Hypothesis 2:** More accurate flow methods will yield better action recognition methods, or in other words, *the accuracy of optical flow is correlated with accuracy of action recognition*. Here we test this hypothesis by using several optical flow methods as input to a baseline action recognition system and recording their classification accuracy. We also measure their accuracy on standard optical flow benchmarks [6, 15] and compute the correlation between these two measurements. Surprisingly, we observe that the EPE of a flow method is poorly correlated with the classification accuracy of a system that uses it as input. We further observe that certain properties of the estimated flow field are more correlated with classification accuracy than others. In particular, the accuracy at motion boundaries and the correct estimation of small displacements are correlated with performance. Overall, this suggests that the prevailing approach of combining optical flow, optimized for EPE, with classification systems may not be optimal.

**Hypothesis 3:** *Optical flow is the best motion representation for action recognition.* Optical flow is often formulated as the problem of estimating the 2D projection of the true 3D motion of the world. In the light of the findings above one may wonder if there exists a criterion by which to choose a particular optical flow method for an action recognition system? To answer this question, we combine the two modules, optical flow estimation and action recognition, in an end-to-end trainable network. We fine-tune the optical flow network by optimizing the final classification accuracy of the full model rather than using the traditional EPE loss, which encourages good performance in classical optical flow benchmarks. In our experiments, we consider two different deep learning-based flow models (FlowNet [8] and SpyNet [24]). After end-to-end training, the networks still compute a representation that looks like optical flow but that improves action recognition performance compared to the original flow. In the case of FlowNet, performance on classical flow benchmarks decreases while, for SpyNet, there is no significant change. Finally, we compare the original optical flow to the new motion representation qualitatively and quantitatively, and observe that the new motion fields are most different on the human body and near the boundary of the body; that is the networks appear to focus on the human and their motion.

The conclusions drawn by this paper shed new light on the way optical flow and video recognition modules are used today. In summary, our conclusions are:

- Optical flow is useful for action recognition as it is invariant to appearance, even when temporal coherence is not maintained.

- EPE of current methods is not well correlated with action recognition accuracy, and thus improving EPE may not improve action recognition.

- Flow accuracy at boundaries and for small displacements matters most for action recognition.

- Training optical flow to minimize classification error instead of EPE improves action recognition results.

- Optical flow that is learned to minimize classification differs on the human body and near boundaries.

These experiments suggest that classical optical flow, developed in isolation, may not be the best representation for video and action classification. This argues against the modular approach to vision and in favor of a more integrated approach.

## 2. Related Work

**Optical flow estimation.** The field of optical flow has made significant progress by focusing on improving numerical accuracy on standard benchmarks. Flow is seen as a source of input for tracking, video segmentation, depth estimation, frame interpolation, and many other problems. It is assumed that optimizing for low EPE will produce flow that is widely useful for many tasks. EPE, however, is just one possible measure of accuracy and others have been used in the literature, such as angular error [3] or frame interpolation error [2].

While there is extensive research on optical flow, here we focus on methods that use deep learning because these can be trained end-to-end on different tasks with different loss functions. Applying learning to the optical flow problem has been hard because there is limited training data with ground truth flow. Early approaches used synthetic data and broke the problem into pieces to make learning possible with limited data [26, 31].

The first end-to-end trainable deep convolutional network is FlowNet [8], which is trained on synthetic data to minimize EPE. The results, however, are not on par with top methods, mostly due to inaccuracies for very large or very small displacements. To address this issue, Ranjan and Black [24] propose SpyNet, a combination of the traditional pyramid approach and convolutional networks. This reduces the size of the network, and improves numerical results. The top performing learning-based flow method is FlowNet2 [13]. Here the authors stack multiple FlowNet modules [8] together, train a separate network to focus on small displacements, and experiment carefully with training schedules. The network is complex but produces competitive, though not top, results on the Sintel benchmark [6] while being near the best monocular methods for KITTI 2015 [21]. The complexity makes it difficult to train as part of a system because each component is trained separately and accuracy is sensitive to different training schedules.

Here we retrain both FlowNet and SpyNet on the action recognition problem and find consistent results. In both cases, training on the recognition loss results in a motion representation that is better for action recognition.

**Motion and action recognition.** For human action recognition, there is significant evidence to suggest the importance of motion as a cue. Johansson [17] argues for separating form from motion and demonstrates that 10-12 dots located at the joints of a moving person are sufficient for human observers to recognize a range of human actions. Bobick and Davis [4] show that humans and algorithms can recognize actions in very low resolution video sequences where, in any given frame, the action is unrecognizable. These results would suggest that optical flow can and does play an important role in action recognition. As a counterpoint, however, Koenderink et al. [19] show that movies can undergo significant spatial and temporal distortions yet human observers can still recognize the actions in them. Even with frames that are scrambled in space and time, people can robustly recognize human actions.

More computationally, Jhuang et al. [16] ask what it is about the flow field that is important for human action recognition. They note that "the motion boundaries around a person's body contour seem to contain information for action recognition that is as important as the optical flow within the region of the body." This hints at the idea that flow may be useful for not only extracting *motion*, as suggested by Johansson, but also *form*, that is the shape of the objects performing the action.

**Learning action recognition.** Top performing deep learning methods for action recognition use one or more architecture styles in order to capture temporal structure: two-streams, 3D convolutions or recurrent neural networks (RNNs). Two-stream architectures [28, 10, 33, 34] typically contain two parallel networks adapted from the image recognition literature, such as VGG [29] or Inception [14]. One network, often referred to as the spatial stream, takes as input one or more images. The other network referred to as the temporal stream takes as input one or more flow fields. At test time, the predictions of each of these networks are combined to output the final label. Many different variants of this architecture have been explored and improved upon the basic structure [9, 11, 27]. Another common style is the use of 3D convolutions [18, 32]. These architectures operate on the video as a volume, and thus they may not use optical flow, as the temporal structure is expected to be captured implicitly by the convolution in the third dimension. Carreira and Zisserman [7] propose I3D, which combines elements of these two styles. Finally, RNNs architectures like long short-term memory networks (LSTMs) [23], have also shown successful results.

The action recognition literature is very broad and a thorough review is beyond our scope. Here we focus on two-stream networks because they tend to have top performance and because they use optical flow as input, which makes them most relevant to us. Since flow is computed from the image sequence, one could argue that a network trained to do action recognition could learn to compute flow if it is useful, making the explicit computation of optical flow unnecessary. Of course, whether this works on not may de-pend on the network architecture. In fact, ablative studies confirm [33, 34] that the use of computed optical flow (*i.e.*, using the temporal stream) improves classification over using only the images (*i.e.*, using only the spatial stream). This indicates the difficulty of learning the task end-to-end. As noted for FlowNet2, even supervised training of a flow network can be tricky.

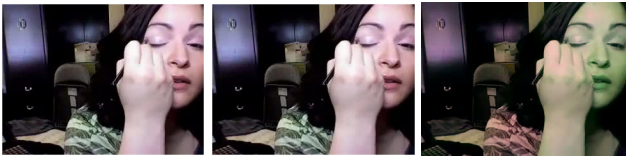

Figure 2. **Samples of modified input videos.** Left: Original image. Middle: Modified image, by sampling a different colormap, referred as "altered colormap". Right: Modified image by scaling channel values, referred as "shift color".

| Input | Accuracy |
|---|---|
| Flow | 86.85% |
| RGB | 85.43% |
| Flow (shuffled flow fields) | 78.64% |
| Flow (shuffled images) | 59.55% |
| Flow (altered colormap) | 84.30% |
| RGB (altered colormap) | 34.23% |
| Flow (shift color) | 85.71% |
| RGB (shift color) | 62.65% |

Table 1. **Action recognition accuracy with modified inputs**. Removing temporal structure by shuffling ("shuffled flow fields" and "shuffled images") affects the performance of the network, but it is still far from chance, suggesting that the benefit of the flow is not just that it represents motion. Modifying the appearance of the input images ("altered colormap" and "shift color") hurts the performance when using only the images but leaves the performance unchanged when using flow.

One of the most widely used two-stream networks is Temporal Segment Networks (TSN) [34]. Each of the streams is composed by an "inception with batch normalization" network [14]. Small snippets of are sampled throughout the video and predictions on each of these subsets are aggregated in the end. Here we work with this model as our baseline.

The most related work to ours is the unpublished (arXiv) ActionFlowNet [22]. Here the authors also explore the idea of using action classification accuracy as a loss to fine-tune an optical flow network, obtaining similar numerical improvements. However, that work does not study in detail the interaction of the two modules (flow and action recognition) but focuses more on the general numerical improvement of the action recognition task, adding, for example, a multi-task loss that combines EPE and action recognition. Instead, here we focus on understanding why flow is useful for action recognition, what makes a flow method better or worse for action recognition and how flow changes after fine-tuning on action recognition. In particular, we show experiments varying the temporal structure of the flow and the appearance of the input image (Sec. 3), provide a comprehensive study of the correlation of EPE and action recognition accuracy with 6 different flow methods (Sec. 4) and analyze how flow fields change after fine-tuning on the action recognition task (Sec. 5).

## 3. Why Use Optical Flow as Input for Video Classification?

Although it may seem intuitive to use explicit motion estimation as input for a task involving video, one could argue that using motion is not essential. Some possible arguments are that categories in current datasets can be identified from a single frame, and more broadly many objects and actions in the visual world can be identified from a single frame. On the other hand, there is evidence that motion contains useful information for recognition of humans, as shown in Johansson's work [17]. One could also argue that the raw video stream implicitly contains the information present in optical flow and, if a network needs this, it can compute it. Despite these arguments, it has been shown in the action recognition literature that using optical flow as input in addition to the raw images improves classification performance [33]. Thus why is optical flow useful for action recognition?

The intuitive answer is that, as in the case of Johansson's work [17], motion trajectories contain enough information to recognize actions. We test this hypothesis by removing temporal coherence in the input video and measuring how this affects recognition accuracy. We remove temporal coherence by shuffling the flow fields randomly in the temporal dimension.

We choose the state-of-the-art Temporal Segment Network (TSN) [34] for our experiments as it is widely used and exploits a two-stream network, which lets us experiment with the two input modalities separately (images used for the spatial stream and flow used for the temporal stream). We evaluate this network on the UCF101 dataset [30]. The temporal stream of the TSN processes the optical flow in blocks of 5 flow fields, corresponding to 6 image frames. We perform the temporal shuffling at test time in the input videos, within each block.[1] The ordering of the temporal shuffle is random within each block. This breaks temporal coherence.

The results are shown in Table 1. While the recognition accuracy decreases when the trajectories are disrupted, the overall accuracy is still surprisingly high (78.64%, table entry "Flow (shuffled flow fields)"), which suggests that coherent temporal trajectories are not the most important feature for recognition, at least in this particular set up and considering this particular architecture.

---

[1] We do not experiment shuffling the frames and using them as input to the spatial stream because it takes a single frame at a time, thus the temporal ordering is already discarded.

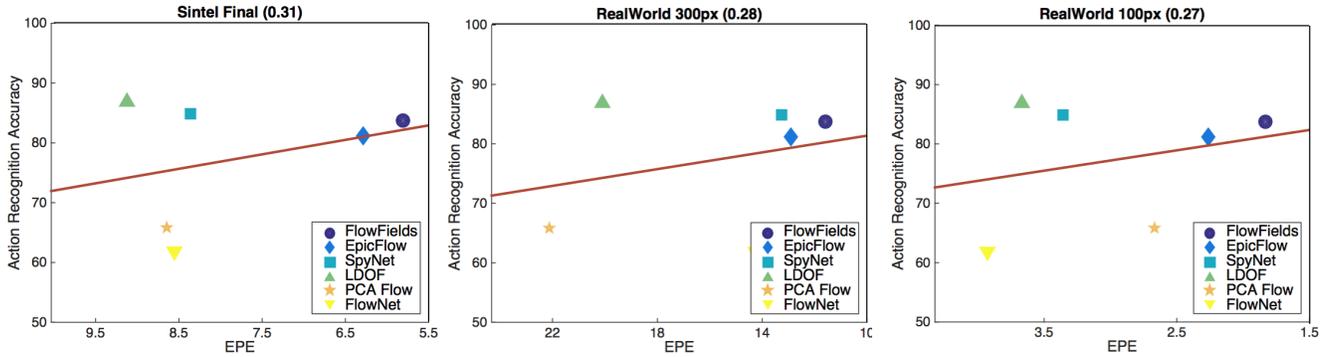

Figure 3. **Action recognition accuracy vs. EPE.** We observe that action recognition accuracy is not very highly correlated in general with EPE on Sintel and Real-World Scenes.

There is still, however, some temporal information encoded in the shuffled flow fields since each of these is computed from two temporally adjacent frames. We remove this temporal information completely by applying the same shuffling scheme to the original input RGB frames before computing the optical flow. At this point, the flow does not correspond to the physical motion in general, but it still captures the shapes of moving objects. The results are also shown in Table 1. This time the recognition accuracy decreases to 59.5% (table entry "Flow (shuffled images)"). While this is a substantial drop, it is still well above chance-level accuracy which is $\sim 1\%$, suggesting that much of the value of the optical flow is not even in the motion per se.

We argue that instead much of the value of optical flow is that it is invariant to the appearance of the input images, resulting in a simpler learning problem as less variance needs to be captured by the action recognition network. To illustrate this, let us consider two videos (one in the training set and one in the test set) that are identical except that the actor is wearing differently colored clothing. If we use the raw images as input to the recognition method, the representation of these two videos will be different and therefore it will be more difficult to learn from one and generalize to the other. However, if we first input such videos to a flow method, the flow of both videos will be very similar, making it easier for the recognition system to learn from one and generalize to the other.

We test this intuition by altering the colormap of the UCF101 and observing the change in accuracy. We do this by converting the RGB images to grayscale and then converting them back to color with a randomly sampled colormap (*e.g.*, jet, brg, hsv). An example of this mapping in shown in Fig. 2. Results are shown in Table 1. We observe that using the modified images as input the accuracy drops by 50% (from the original result in entry "RGB" to "RGB (altered colormap)"), while using only the flow (computed from the modified images) the accuracy decreases only marginally (table entry "Flow (altered colormap)"). While this alteration of the appearance could resemble the case of two videos where the person is wearing different color clothing, it may result in a very large change over the entire frame. To explore what happens with a smaller change of appearance, we also alter the input frames by scaling the value of each channel by a random coefficient between 0.3 and 1. The result of this change is much less noticeable in the image, but it still impacts the performance of the network using only images by 20%, while it leaves the accuracy of the network using optical flow almost the same.

While these results are not surprising, they confirm with numerical evidence the intuition that the temporal structure is not responsible for the success of optical flow in many action recognition applications. Instead, the invariance to appearance of the representation is key.

## 4. Is Optical Flow Accuracy Correlated with Action Recognition Accuracy?

Progress in the optical flow and the action recognition communities has been mostly independent. However, it remains unclear if these separate advances lead to a common goal. In other words, does a better optical flow method (as judged by the optical flow community) improve action recognition accuracy? In this section we test whether EPE, which is the standard metric for optical flow, is correlated with classification accuracy.

We use the temporal stream of the TSN network [34], and create different versions, fine-tuning each of them with a different optical flow method. These optical flow methods are chosen to cover a wide range of accuracies and styles (deep learning based, variational, etc). We choose FlowFields [1], EpicFlow [25], LDOF [5], PCAFlow [35], FlowNet [8] and Spynet [24]. We take the TSN [34] pre-trained using DenseFlow as initialization and fine-tune it with each flow method separately. To validate the fairness of the initialization scheme, we also conducted experiments with the initial model trained on ImageNet. Both initialization schemes lead to the same ranking of flow methods on action recognition. The results are displayed in Fig. 3.

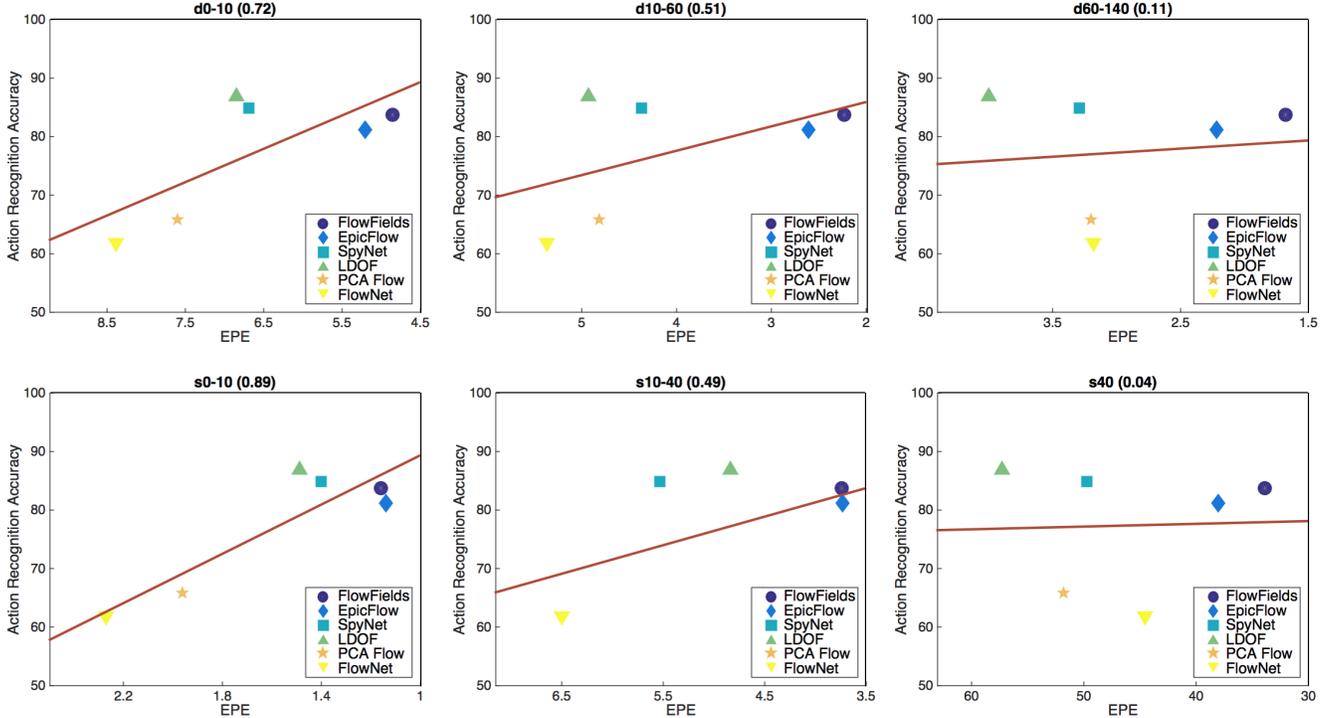

Figure 4. **Action recognition accuracy vs. EPE in specific regions of the scene**. The regions are categorized according to distance to the boundary ("d0-10", "d10-60", and "d60-140" are pixels less than 10p away from a motion boundary, 10-60p and 60-140p, respectively), and according to speed ("s0-10", "s10-40" and "s40+" are pixels that move less than 10p, between 10 and 40, and more than 40, respectively). We observe that error at small displacements and close to the boundary the correlation with recognition accuracy is higher.

Since there is no ground truth flow on UCF101, we measure the flow methods according to their performance on standard benchmarks. For a fair evaluation, we consider both the synthetic MPI Sintel dataset [6] and the recently introduced, more realistic Real-World Scenes dataset of Janai et al. [15]. Figure 3 shows the end-point-error (EPE) of each flow method in terms of action recognition accuracy on the UCF101 dataset, along with the correlation score. Note that the EPE along the horizontal axis is in decreasing order as smaller EPE indicates better flow in general. Since the FlowNet model fine-tuned on MPI Sintel is not released, we evaluate FlowNet ourselves by submitting it to the Sintel server. For the Real-World Scenes, we present the EPE value on a clean version with 100px flow magnitude without motion, and also a hard version with 300px flow magnitude in addition to motion blur.

Figure 3 shows that smaller EPE does not correlate strongly with better action recognition. For example, LDOF has relatively large EPE on both Sintel and Real-World Scenes, but it produces the best performance when used as input for action recognition. Although the EPE rankings of these flow methods vary between different datasets, the correlation between flow error and action recognition accuracy is relatively weak.

This finding may be attributed to the fact that EPE is computed and averaged over the entire image. However, not all flow vectors contain the same amount of information for action recognition. For example boundaries contain information about shape, which is useful for recognition but large camera motions in the background may not be very informative. We analyze the effect of different regions of the scene on recognition using Sintel's additional annotations. At each pixel, Sintel contains a label about the speed of the flow (small displacements 0-10 pixels, medium 10-40p, and large 40+p). Sintel also contains a label about the distance to motion boundary (0-10p, 10-60p, 60-140p). Figure 4 shows the correlation between the error in each of these regions of the scene and the action recognition accuracy. While there is not a very strong correlation in any of these regions, the most relevant for recognition are small displacements (where the Pearson correlation coefficient is $\rho = 0.89$). This is reasonable since small displacements are common in human actions (eg. knitting, applying eye make up, etc) such as those present in UCF101. The second most important regions are boundaries ($\rho = 0.72$), which inform about shape. Flow accuracy of the tested methods on large motions is uncorrelated with action recognition performance. This could be because all flow methods are relatively poor at estimating large motions. If they are effectively estimating noise, then we would expect this to be uncorrelated. With our experiments it is not possible to decouple whether this lack of correlation is due to the poor flow or to large motions not being important for human action recognition.

| Model | EPE Loss | Action Loss |
|---|---|---|
| FlowNet (3 Sn.) | 45.97% | **50.51**% |
| FlowNet (25 Sn.) | 45.96% | **50.98**% |
| FlowNet (25 Sn. + Data Aug.) | 56.86% | **59.41**% |
| SpyNet (3 Sn.) | 68.75% | **70.45**% |
| SpyNet (25 Sn.) | 70.26% | **71.50**% |
| SpyNet (25 Sn. + Data Aug.) | 80.91% | **81.47**% |

Table 2. **Action recognition accuracy using optical flow trained on EPE vs. trained on action recognition accuracy.** We test using multiple evaluation schemes. In "3 Sn." we take 3 snippets of 5 flow frames, and average the predictions; in "25 Sn." the same with 25 snippets; and in "25 Sn. + Data Aug." the same with the over-cropping data augmentation process of TSN. Learning to compute flow for action recognition improves recognition accuracy, across all evaluation schemes and for both flow networks.

This analysis is useful in order to select an optical flow method from an existing repertoire for action recognition. However, it also leads to the question – while improving EPE may lead to progress on action recognition, could there be a better and more direct metric for the task?

## 5. Are There Better Motion Representations for Action Recognition than Optical Flow?

Current approaches that use motion cues for action recognition follow a sequential procedure: first they compute flow and then use it as input to a classification network (for example, TSN uses DenseFlow). The assumption of this approach is that accurate flow (measured by EPE) is useful for action recognition. However, our experiment of Sec. 4 shows that this correlation is weak. Thus can we instead use action recognition accuracy as an optimization criterion? In this section we train the optical flow network to learn a better motion representation directly from the high level task of action recognition.

We use TSN as the recognition network, and experiment with both SpyNet and FlowNet as optical flow modules. We start from the generic flow models and fine-tune them using the action recognition loss back-propagated through the TSN. We use the TSN models fined-tuned to each of the flow methods in the experiment of the previous section, and keep them fixed to observe how the optical flow changes.[4] We follow the training scheme of TSN with 3 blocks of 6 consecutive frames (5 flow fields) randomly sampled. We use a small learning rate of $1e-7$, as in the original FlowNet.

---
[3]https://distill.pub/2016/deconv-checkerboard/
[4]The SpyNet model is slightly different in this section than the previous

| Method | EPE all | EPE matched | EPE unmatched |
|---|---|---|---|
| FlowNet (EPE loss) | 8.552 | 5.053 | 37.051 |
| FlowNet (Action loss) | 8.654 | 5.149 | 37.185 |
| SpyNet (EPE loss) | 10.715 | 6.377 | 46.046 |
| SpyNet (Action loss) | 10.719 | 6.400 | 45.906 |

Table 3. **Evaluation in Sintel of FlowNet and SpyNet, trained on EPE and trained on action recognition accuracy.** EPE slightly increases after the networks are fine-tuned for action recognition.

**Experiments on action recognition and optical flow.** We find that learning to compute flow to recognize actions improves the motion features and thus the action recognition accuracy. Results are shown in Table 2. In this table we present different evaluation schemes. Using the evaluation scheme of TSN (named "25 Sn. + Data Aug." in the table) we observe that FlowNet improves by almost 3% and SpyNet by 0.5%. In this evaluation scheme, 25 snippets (each snippet being a 5 flow field window) are sampled and evaluated and their predictions averaged. In addition, there is data augmentation at test time, where each snippet is "overcropped"[5] 10 times. While this evaluation scheme produces higher accuracy values, it is expensive at test time. Therefore we also show the results of evaluating without data augmentation (named "25 Sn." in the table), where absolute accuracy decreases but we observe the same pattern, of both FlowNet and SpyNet improving their accuracy after learning using the action recognition loss. Finally, we use a light evaluation scheme with only 3 snippets ("3 Sn."). We observe that both flow methods improve their accuracy even more (4% in the case of FlowNet and 2% in the case of SpyNet) in this evaluation scheme that takes in less image evidence, and therefore the quality of the features becomes more apparent. These improvements across all models and evaluation schemes suggest that task-specific flow estimation may be beneficial for solving higher level tasks.

We also evaluate the task-specific models on Sintel and observe a similar or slight increase in EPE (Table 3). This is consistent with the results of the previous section where EPE does not correlate strongly with recognition accuracy.

**Experiments on the statistics of the new motion representation.** What do the learned task-specific optical flow fields represent? We compare the flow fields estimated by

---
one, since we used the end-to-end trainable version.
[5]This overcropping process consists of cropping the top-left corner, top-right corner, center, bottom-left corner and bottom-right corner, and their flipped counterparts.

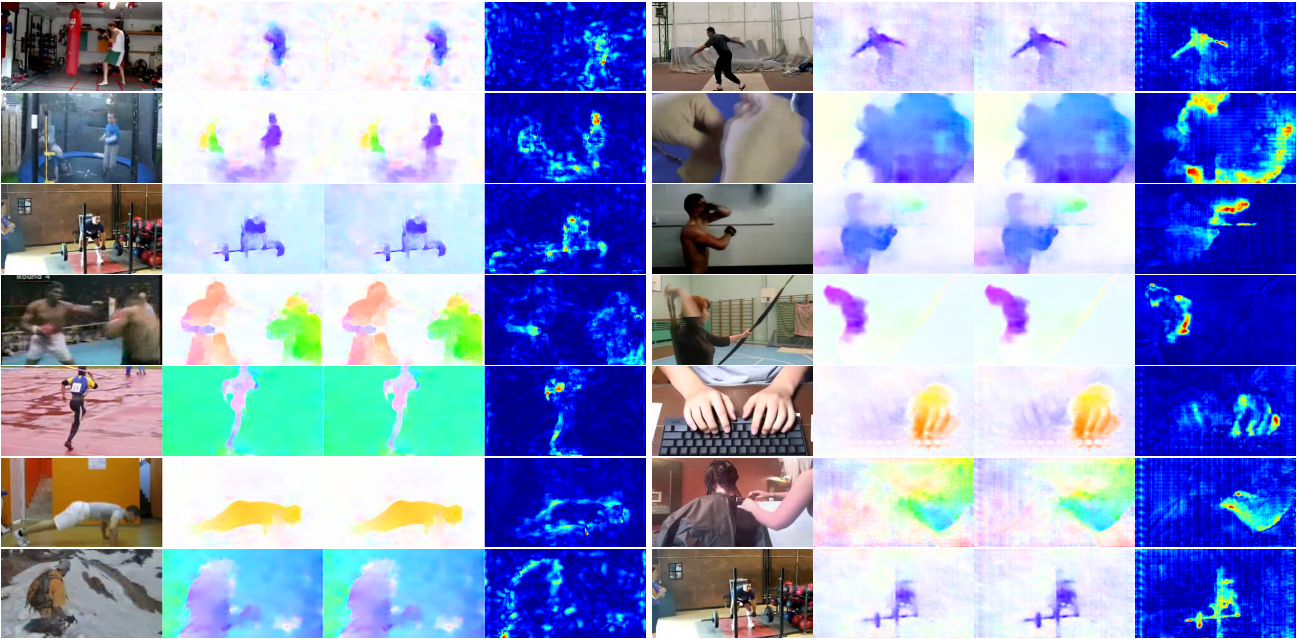

Figure 5. **Comparison of estimated flow fields from a network trained to minimize EPE and the same trained to minimize classification error.** Left: Results from SpyNet. Right: Results from FlowNet. Each series of 4 images represent the first of the input images, the two flow fields, and the Euclidean distance between the two flow vectors at each pixel (red means higher distance, and blue means lower distance). Flow vectors noticeably change more around motion boundaries and where humans are located. In the case of FlowNet we observe the appearance of a checker-board pattern attributed to the upconvolution [3], but we observe that it does not affect the improvement of performance for action recognition. SpyNet does not exhibit the artefact since it does not contain upconvolution.

the model trained on EPE and the model trained on action recognition. The comparisons are shown in Fig. 5. For each pair of flow fields, the figure shows the first image, the two optical flow fields (using an EPE loss and an action recognition loss), and the Euclidean distance between the two. We observe that most changes occur in two very specific regions of the scene: at motion boundaries and where humans are located. This behavior is consistent both in SpyNet and FlowNet. Some examples of this effect are the fingers on the typing right hand, the boundary of the person doing push-ups or the arm of the person doing archery.

To help us quantify this change we use the Mask-RCNN method [12], to estimate the regions where humans are. We compare the average change of flow at each pixel inside versus outside the human mask. The results are shown in Table 4. In both networks pixel values change one order of magnitude more in regions where humans are located than outside. We quantify the change at boundaries by computing the edge of the mask and dilating it to obtain a thickness of 20 pixels, which captures pixels 10p away from the boundary. In Table 4 we observe that pixels at the boundary change an order of magnitude more than pixels elsewhere. This suggests that flow at boundaries and at objects of interest (in this case humans), is most important for recognition.

| Method | Human Pixels | Non-Human Pixels | Boundary Pixels | Non-Bound. Pixels |
|---|---|---|---|---|
| SpyNet | 0.1181 | 0.0169 | 0.1054 | 0.0185 |
| FlowNet | 0.5046 | 0.0721 | 0.4683 | 0.0821 |

Table 4. **Statistics of the new motion representation**, Regions of the scene where the optical flow changes most during training. We observe that in both methods flow changes most at boundaries and on the human regions.

## 6. Conclusion

We presented an analysis of two computer vision building blocks (optical flow and action recognition). These two modules are often used together but their interaction has hardly been analyzed. Through thorough experimentation with one of the state-of-the-art action recognition methods and a wide range of optical flow methods we make a number of observations. Optical flow is useful because it is invariant to appearance, even when the flow vectors are inaccurate. We also observe that the traditional EPE metric is weakly correlated with action recognition accuracy but EPE at boundaries and on small displacements is more relevant for recognition. To compute flow that is better for the task we also learned optical flow to directly minimize action recognition error. This leads to numerical improvements on action recognition accuracy. We observed that these improvements arise from changes in the flow on the human body and near the boundary of the body. We believe that our observations will help optical flow researchers who are interested in applications of optical flow for recognition tasks, as well as action recognition researchers who wish to make better decisions about their motion representations.